\documentclass[conference]{IEEEtran}
\usepackage{times}

\usepackage[numbers]{natbib}
\usepackage{multicol}
\usepackage[bookmarks=true,breaklinks=true]{hyperref}
\usepackage{amssymb}
\usepackage{amsbsy}
\usepackage{graphicx}
\usepackage{subcaption} 
\usepackage[dvipsnames]{xcolor}
\usepackage{soul} % Added by atil for strike out command
\usepackage{authblk}
\begin{document}
% paper title
\title{Sim-to-Real: Learning Agile Locomotion For Quadruped Robots}

\author[1]{Jie Tan}
\author[1]{Tingnan Zhang}
\author[1]{Erwin Coumans}
\author[1]{Atil Iscen}
\author[2]{\\Yunfei Bai}
\author[1]{Danijar Hafner}
\author[3]{Steven Bohez}
\author[1]{Vincent Vanhoucke}
\affil[1]{Google Brain}
\affil[2]{X}
\affil[3]{Google DeepMind}

\maketitle

\begin{abstract}
Designing agile locomotion for quadruped robots often requires extensive expertise and tedious manual tuning. In this paper, we present a system to automate this process by leveraging deep reinforcement learning techniques. Our system can learn quadruped locomotion from scratch using simple reward signals. In addition, users can provide an open loop reference  to guide the learning process when more control over the learned gait is needed. The control policies are learned in a physics simulator and then deployed on real robots. In robotics, policies trained in simulation often do not transfer to the real world. We narrow this reality gap by improving the physics simulator and learning robust policies. We improve the simulation using system identification, developing an accurate actuator model and simulating latency. We learn robust controllers by randomizing the physical environments, adding perturbations and designing a compact observation space. We evaluate our system on two agile locomotion gaits: trotting and galloping. After learning in simulation, a quadruped robot can successfully perform both gaits in the real world.
\end{abstract}

\IEEEpeerreviewmaketitle

\section{Introduction}
Designing agile locomotion for quadruped robots is a long-standing research problem \cite{Raibert:1986:LRB:6152}. This is because it is difficult to control an under-actuated robot performing highly dynamic motion that involve intricate balance. Classical approaches often require extensive experience and tedious manual tuning \cite{pratt1998intuitive,de2017}. Can we automate this process? 

Recently, we have seen tremendous progress in deep reinforcement learning (deep RL) \cite{lillicrap2015continuous, DBLP:journals/corr/SchulmanWDRK17, Duan:2016:BDR:3045390.3045531}. These algorithms can solve locomotion problems from scratch without much human intervention. However, most of these studies are conducted in simulation, and a controller learned in simulation often performs poorly in the real world. This \emph{reality gap} \cite{koos2010crossing, boeing2012leveraging} is caused by model discrepancies between the simulated and the real physical system. Many factors, including unmodeled dynamics, wrong simulation parameters, and numerical errors, contribute to this gap. Even worse, this gap is greatly amplified in locomotion tasks. When a robot performs agile motion with frequent contact changes, the switches of contact situations break the control space into fragmented pieces. Any small model discrepancy can be magnified and generate bifurcated consequences. Overcoming the reality gap is challenging. 

\begin{figure}[!t]
    \centering
    \includegraphics[width=3.4in]{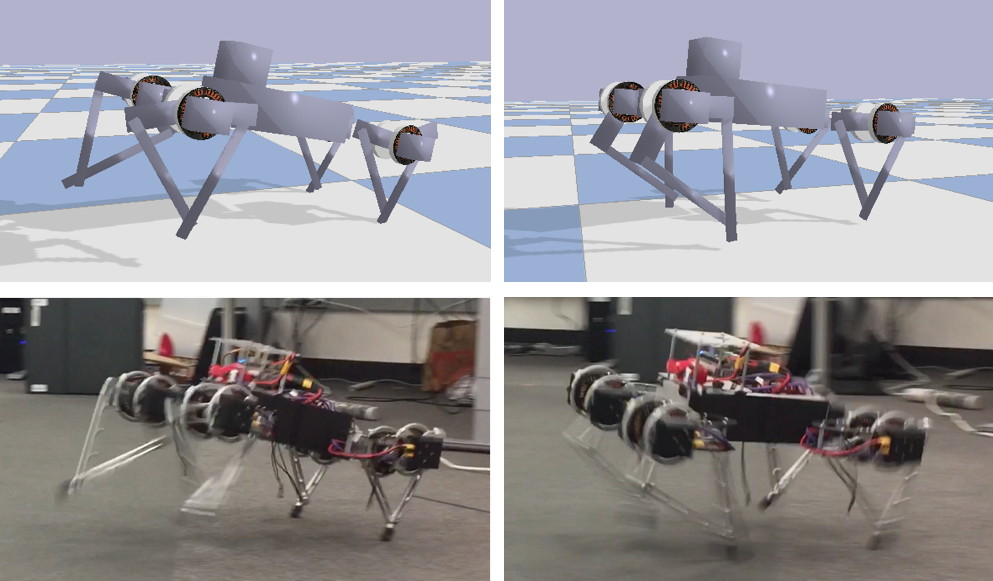}
    \caption{The simulated and the real Minitaurs learned to gallop using deep reinforcement learning.}
    \label{fig:simReal}
\end{figure}

An alternative is to learn the task directly on the physical system. While this has been successfully demonstrated in robotic grasping \cite{levine2016learning}, it is challenging to apply this method to locomotion tasks due to the difficulties of automatically resetting the experiments and continuously collecting data. In addition, every falling during learning can potentially damage the robot. Thus for locomotion tasks, learning in simulation is more appealing because it is faster, cheaper and safer.

In this paper, we present a complete learning system for agile locomotion, in which control policies are learned in simulation and deployed on real robots. There are two main challenges: 1) learning \emph{controllable} locomotion policies; and 2) transferring the policies to the physical system. 

While learning from scratch can lead to better policies than incorporating human guidance \cite{silver2017mastering}, in robotics, having control of the learned policy sometimes is preferred. Our learning system provides users a full spectrum of controllability over the learned policies. The user can choose from letting the system learn completely by itself to specifying an open-loop reference gait as human guidance. Our system will keep the learned gait close to the reference while, at the same time, maintain balance, increase speed and energy efficiency.

To narrow the reality gap, we perform system identification to find the correct simulation parameters. Besides, we improve the fidelity of the physics simulator by adding a faithful actuator model and latency handling. To further narrow the gap, we experiment with three approaches to increase the robustness of the learned controllers: dynamics randomization, perturbation forces, and compact design of observation space.

We evaluate our system on a quadruped robot with two locomotion tasks, trotting and galloping, in Section \ref{sec:evaluation}. We show that with deep RL, highly agile locomotion gaits can emerge automatically. We also demonstrate how users can easily specify the style of locomotion using our system. When comparing with the gaits handcrafted by experts, we find that our learned gaits are more energy efficient at the same running speed. We demonstrate that with an accurate physics simulator and robust control policies, we can successfully deploy policies learned in simulation to the physical system. The main contributions of this paper are:

\begin{enumerate}
    \item We propose a complete learning system for agile locomotion. It provides users a full spectrum (from fully restricted to a user-specified gait to fully learned from scratch) of controllability over the learned policies.
    \item We show that the reality gap can be narrowed by a variety of approaches and conduct comprehensive evaluations on their effectiveness.
    \item We demonstrate that agile locomotion gaits, such as trotting and galloping, can be learned automatically and these gaits can work on robots directly without further training on the physical system.
\end{enumerate}

\section{Related Work}
\subsection {Legged Locomotion Control}
Optimizing controllers \cite{kober2012reinforcement} automatically is an appealing alternative to tedious manual tuning. Popular optimization methods for locomotion control include black-box \cite{icra18-minitaur} and Bayesian optimization \cite{cully2015robots}. Bayesian optimization is often data efficient enough to be applied directly on real robots \cite{calandra2016bayesian, DBLP:conf/corl/AntonovaRA17}. However, it is challenging to scale these methods to high-dimensional control space. For this reason, feature engineering and controller architecture design are needed. 

On the other hand, recent advances in deep RL have significantly reduced the requirement of human expertise \cite{Duan:2016:BDR:3045390.3045531}. We have witnessed intense competitions among deep RL algorithms in the simulated benchmark environments \cite{1606.01540,tassa2018deepmind}. In this paper, we choose to use Proximal Policy Optimization (PPO) \cite{DBLP:journals/corr/SchulmanWDRK17} because it is a stable on-policy method and can be easily parallelized \cite{hafner2017agents}.

Much research has applied reinforcement learning to locomotion tasks \cite{benbrahimbiped,tedrake2004stochastic,endo2005learning,icra04,DBLP:journals/ar/OginoKAAH04}. More recently, Gay et al. \cite{gay2013learning} learned a neural network that modified a Central Pattern Generator controller for stable quadruped locomotion. Levine et al. \cite{2014-cgps} applied guided policy search to learn bipedal locomotion in simulation. Peng et al. \cite{2015-TOG-terrainRL} developed a CACLA-inspired algorithm to control a simulated dog to navigate complex 1D terrains. It was further improved using a mixture of actor-critic experts \cite{2016-TOG-deepRL}. Peng et al. \cite{2017-TOG-deepLoco} learned a hierarchical controller to direct a 3D biped to walk in a simulated environment. Heess et al. \cite{DBLP:journals/corr/HeessTSLMWTEWER17} showed that complex locomotion behaviors, such as running, jumping and crouching, can be learned directly from simple reward signals in rich simulated environments. Sharma and Kitani \cite{Sharma-2018-103569} exploited the cyclic nature of locomotion with phase-parametric policies.  Note that in most of these latest work, learning and evaluation were exclusively performed in simulation. It is not clear whether these learned policies can be safely and successfully deployed on the robots. In this paper, while we also train in simulation, more importantly, we test the policies on the real robot and explore different approaches to narrow the reality gap.

\subsection{Overcoming the Reality Gap}
Reality gap is the major obstacle to applying deep RL in robotics. Neunert et al. \cite{neunert2017off} analyzed potential causes of the reality gap, some of which can be solved by system identification \cite{DBLP:journals/corr/abs-1710-08893}. Li et al. \cite{li2013terradynamics} showed that the transferability of open loop locomotion could be increased if carefully measured physical parameters were used in simulation. These physical parameters can also be optimized by matching the robot behaviors in the simulated and the real world \cite{tan2016simulation, escidoc:2316379}. Bongard et al. \cite{Bongard2006} used the actuation-sensation relationship to build and refine a simulation through continuous self-modeling, and later used this model to plan forward locomotion. Ha et al. \cite{ha2015reducing} used Gaussian Processes, a non-parametric model, to minimize the error between simulation and real physics. Yu et al. \cite{DBLP:journals/corr/YuLT17} performed online system identification explicitly on physical parameters, while Peng et al. \cite{peng2017sim} embedded system identification implicitly into a recurrent policy. In addition to system identification, we identify that inaccurate actuator models and lack of latency modeling are two major causes of the reality gap. We improve our simulator with new models.

A robust controller is more likely to be transferred to the real world. Robustness can be improved
by injecting noise \cite{jakobi1995noise}, perturbing the simulated robot \cite{pinto2017robust}, leveraging multiple simulators \cite{boeing2012leveraging}, using domain randomization \cite{tobin2017domain} and dynamics randomization \cite{mordatch2015ensemble,rajeswaran2016epopt,peng2017sim}. Although not explicitly using real-world data for training, these methods have been shown effective to increase the success rate of sim-to-real transfer.

Another way to cross the reality gap is to learn from both simulation and real-world data. A policy can be pre-trained in simulation and then fine-tuned on robots \cite{DBLP:journals/corr/RusuVRHPH16}. Hanna and Stone \cite{AAAI17-Hanna} adapted the policy using a learned action space transformation from simulation to the real world. In visuomotor learning, domain adaptation were applied at feature level \cite{DBLP:journals/corr/TzengDHFPLSD15,DBLP:journals/corr/abs-1710-06422} or pixel level \cite{bousmalis2017using} to transfer the controller. Bousmalis et al. \cite{bousmalis2017using} reduced the real world data requirement by training a generator network that converts simulated images to real images. While some of these methods were successfully demonstrated in robotic grasping, it is challenging to apply them to locomotion due to the difficulties to continuously and safely collect enough real world data. Furthermore, we need to narrow the reality gap of dynamics rather than perception.

\section{Robot Platform and Physics Simulation}
\label{sec:sim}

\begin{figure}[!t]
    \centering
    \includegraphics[width=3.4in]{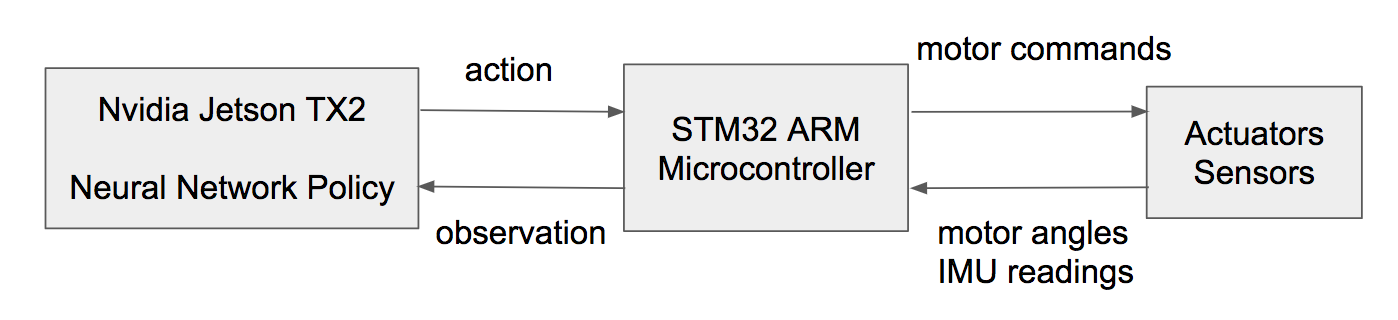}
    \caption{The customized hardware architecture enables the Minitaur to perform deep neural network inference.}
    \label{fig:hardware}
\end{figure}

Our robot platform is the Minitaur from Ghost Robotics (Figure \ref{fig:simReal} bottom), a quadruped robot with eight direct-drive actuators \cite{7403902}. Each leg is controlled by two actuators that allow it to move in the sagittal plane. The motors can be actuated through position control or through a Pulse Width Modulation (PWM) signal. The Minitaur is equipped with motor encoders that measure the motor angles and an IMU that measures the orientation and the angular velocity of its base. An STM32 ARM microcontroller sends commands to actuators, receives sensor readings and can perform simple computations. However, this microcontroller is not powerful enough to execute neural network policies learned from deep RL. For this reason, we installed an Nvidia Jetson TX2 to perform neural network inference. The TX2 is interfaced with the microcontroller through UART communication. At every control step, the sensor measurements are collected at the microcontroller and sent back to the TX2, where they are fed into a neural network policy to decide the actions to take. These actions are then transmitted to the microcontroller and executed by the actuators (Figure~\ref{fig:hardware}). Since the TX2 does not run a real time operating system, the control loop runs at variable control frequencies of approximately 150-200Hz.

We build a physics simulation of the Minitaur (Figure \ref{fig:simReal} top) using PyBullet \cite{coumans2017}, a Python module that extends the Bullet Physics Engine with robotics and machine learning capabilities. Bullet solves the equations of motion for articulated rigid bodies in generalized coordinates while simultaneously satisfying physical constraints including contact, joint limits and actuator models. Then the state of the system is numerically integrated over time using a semi-implicit scheme.

\section{Learning Locomotion Controllers}
\subsection{Background} 

We formulate locomotion control as a Partially Observable Markov Decision Process (POMDP) and solve it using a policy gradient method. An MDP is a tuple $(S, A, r, D, P_{sas'}, \gamma)$, where $S$ is the state space; $A$ is the action space; $r$ is the reward function; $D$ is the distribution of initial states $s_0$, $P_{sas'}$ is the transition probability; and $\gamma\in[0, 1]$ is the discount factor. Our problem is partially observable because certain states such as the position of the Minitaur's base and the foot contact forces are not accessible due to lack of corresponding sensors. At every control step, a partial observation $o \in O$, rather than a complete state $s \in S$, is observed. Reinforcement learning optimizes a policy $\pi: O \mapsto A$ that maximizes the expected return (accumulated rewards) $R$.
\begin{equation}
    \pi^* = arg\,max_\pi E_{s_0\sim D}[R_\pi(s_0)]
\end{equation}

\subsection{Observation and Action Space} 
\label{sec:obs}
In our problem, the observations include the roll, pitch, and the angular velocities of the base along these two axes, and the eight motor angles. Note that we do not include all available sensor measurements in the observation space. For example, the IMU also provides the yaw of the base. We exclude it because the measurement drifts quickly. The motor velocities can also be calculated but can be noisy. We find that our observation space is sufficient to learn the tasks demonstrated in this paper. More importantly, a compact observation space helps to transfer the policy to the real robot. More analysis on this is presented in Section \ref{sec:evaluation}.

\begin{figure}[!b]
    \centering
    \includegraphics[width=2.8in]{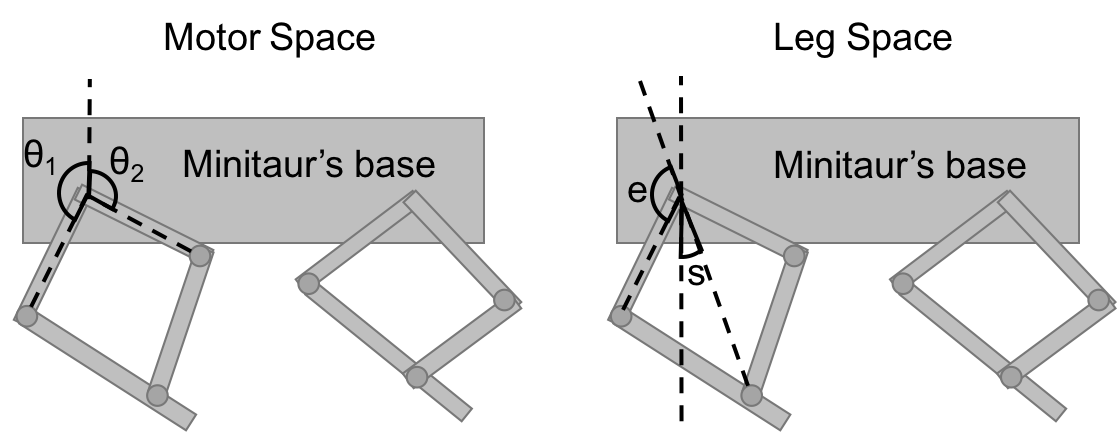}
    \caption{Representation of the leg pose in motor space and leg space. Extension (e) sets the length of the leg by rotating both motors in opposite directions while swing (s) sets the overall rotation of the leg by rotating both motors in same direction.}
    \label{fig:leg_model}
\end{figure}

When designing the action space, we choose to use the position control mode of the actuators for safety and ease of learning \cite{2017-SCA-action}. The actions consist of the desired pose of each leg in the \emph{leg space} \cite{7403902, icra18-minitaur}. The pose of each leg is decomposed into the swing and the extension components $(s, e)$ (Figure \ref{fig:leg_model}). They are mapped into the \emph{motor space} as
\begin{eqnarray}
\nonumber
\theta_1 &= e + s \\
\nonumber
\theta_2 &= e - s
\end{eqnarray}
where $\theta_1$ and $\theta_2$ are the angles of the two motors controlling the same leg; $s$ and $e$ are the swing and the extension components in the leg space. 

An alternative action space is the eight desired motor angles. However, in this motor space, many configurations are invalid due to self collisions between body parts. This results in an action space where valid actions are scattered nonconvex regions, which significantly increases the difficulty of learning. In contrast, in the leg space, we can easily set a rectangle bound that prunes out all the invalid actions while still covering most of the valid configurations.

\subsection{Reward Function}

We design a reward function to encourage faster forward running speed and penalize high energy consumption.
\begin{equation}
r=(\mathbf{p}_{n}-\mathbf{p}_{n-1})\cdot\mathbf{d} - w \Delta t |\boldsymbol{\tau}_n \cdot \dot{\mathbf{q}}_n|
\label{eq:reward}
\end{equation}
where $\mathbf{p}_{n}$ and $\mathbf{p}_{n-1}$ are the positions of the Minitaur's base at the current and the previous time step respectively; $\mathbf{d}$ is the desired running direction; $\Delta t$ is the time step; $\boldsymbol{\tau}$ are the motor torques and $\dot{\mathbf{q}}$ are the motor velocities. The first term measures the running distance towards the desired direction and the second term measures the energy expenditure. $w$ is the weight that balances these two terms. Since the learning algorithm is robust to a wide range of $w$, we do not tune it and use $w=0.008$ in all our experiments. During training, the rewards are accumulated at each episode. An episode terminates after 1000 steps or when the simulated Minitaur loses balance: its base tilts more than 0.5 radians.

\subsection{Policy Representation} Although learning from scratch can eliminate the need of human expertise, and sometimes achieve better performance, having control of the learned policies is important for robotic applications. For example, we may want to specify details of a gait (e.g. style or ground clearance). For this reason, we decouple the locomotion controller into two parts, an open loop component that allows a user to provide reference trajectories and a feedback component that adjusts the leg poses on top of the reference based on the observations.
\begin{equation}
\mathbf{a}(t, \mathbf{o}) = \bar{\mathbf{a}}(t) + \pi(\mathbf{o})
\label{eq:policy}
\end{equation}
where $\bar{\mathbf{a}}(t)$ is the open loop component, which is typically a periodic signal, and $\pi(\mathbf{o})$ is the feedback component. In this way, users can easily express the desired gait using an open loop signal and learning will figure out the rest, such as the balance control, which is tedious to design manually. 

This hybrid policy (eq. (\ref{eq:policy})) is a general formulation that gives users a full spectrum of controllability. It can be varied continuously from fully user-specified to entirely learned from scratch. If we want to use a user-specified policy, we can set both the lower and the upper bounds of $\pi(\mathbf{o})$ to be zero. If we want a policy that is learned from scratch, we can set $\bar{\mathbf{a}}(t)=0$ and give the feedback component $\pi(\mathbf{o})$ a wide output range. By varying the open loop signal and the output bound of the feedback component, we can decide how much user control is applied to the system. In Section \ref{sec:evaluation}, we will illustrate two examples, learning to gallop from scratch and learning to trot with a user provided reference.

We represent the feedback component $\pi$ with a neural network and solve the above POMDP using Proximal Policy Optimization \cite{DBLP:journals/corr/SchulmanWDRK17}. The neural network has two fully-connected hidden layers. Its size is determined via hyperparameter search. Refer to Section \ref{sec:evaluation} for more details. 

\section{Narrowing the Reality Gap}
Due to the reality gap, robotic controllers learned in simulation usually do not perform well in the real environments. We propose two approaches to narrow the gap: improving simulation fidelity and learning robust controllers.

\subsection{Improving Simulation Fidelity}
\label{sec:improve_sim}
Since the reality gap is caused by model discrepancies between the simulation and the real dynamics, a direct way to narrow it is to improve the simulation. We first create an accurate Unified Robot Description Format (URDF) \cite{urdf} file for the simulated Minitaur. We disassemble a Minitaur\footnote{For robots that are difficult to dissemble, traditional system identification methods could be applied.}, measure the dimension, weigh the mass, find the center of mass of each link and incorporate this information into the URDF file. Measuring inertia is difficult. Instead, we estimate it for each link given its shape and mass, assuming uniform density. We also design experiments to measure motor frictions \cite{7403902}. In addition to system identification, we augment the simulator with a more faithful actuator model and latency handling.

\paragraph{Actuator Model} We use position control to actuate the motors of the Minitaur. Bullet also provides position control for simulated motors. In its implementation, one constraint $e_{n+1}=0$ is formulated for each motor where $e_{n+1}$ is an error at the end of current time step. The error is defined as 
\begin{equation}
e_{n+1}=k_p(\bar{q}-q_{n+1}) + k_d(\bar{\dot{q}}-\dot{q}_{n+1})
\label{eq:bullet_motor}
\end{equation}
where $\bar{q}$ and $\bar{\dot{q}}$ are the desired motor angle and velocity, $q_{n+1}$ and $\dot{q}_{n+1}$ are the motor angle and velocity at the end of current time step, $k_p$ is the proportional gain and $k_d$ is the derivative gain. Despite its similarity to the Proportional-Derivative (PD) servo, a key difference is that eq. (\ref{eq:bullet_motor}) guarantees that the motor angle and velocity at the end of the time step satisfy the constraint while PD servo uses the current motor angle and velocity to decide how to actuate the motors. As a result, if large gains are used, the motors can remain stable in simulation but oscillate in reality. 

To eliminate the model discrepancy for actuators, we develop an actuator model according to the dynamics of an ideal DC motor. Given a PWM signal, the torque of the motor is 
\begin{eqnarray}
\label{eq:dc_motor}
\nonumber
\tau & = & K_t I \\ \label{eq:current}
I & = & \frac{V_{\textrm{pwm}}-V_{\textrm{emf}}}{R}\\ \label{eq:emf}
V_{\textrm{emf}} & = & K_t\dot{q} 
\end{eqnarray}
where $I$ is the armature current, $K_t$ is the torque constant or back electromotive force (EMF) constant, $V_{\textrm{pwm}}$ is the supplied voltage which is modulated by the PWM signal, $V_{\textrm{emf}}$ is the back EMF voltage, and $R$ is the armature resistance. The parameters $K_t$ and $R$ are provided in the actuator specification. 

Using the above model, we observe that the real Minitaur often sinks to its feet or cannot lift them while the same controller works fine in simulation. This is because the linear torque-current relation only holds for ideal motors. In reality, the torque saturates as the current increases. For this reason, we construct a piece-wise linear function to characterize this nonlinear torque-current relation \cite{DBLP:journals/corr/DeK15}. In simulation, once the current is computed from PWM (eq. (\ref{eq:current}) and (\ref{eq:emf})), we use this piece-wise function to look up the corresponding torque. 

In position control, PWM is controlled by a PD servo.
\begin{equation}
V_{\textrm{pwm}}=V(k_p(\bar{q}-q_{n}) + k_d(\bar{\dot{q}}-\dot{q}_{n}))
\label{eq:pd}
\end{equation}
where $V$ is the battery voltage. Note that we use the angle and velocity at the current time step. In addition, we set the target velocity $\bar{\dot{q}}=0$ after consulting Ghost Robotics's PD implementation on the microcontroller.

\begin{figure}
    \centering
    \includegraphics[width=2.4in]{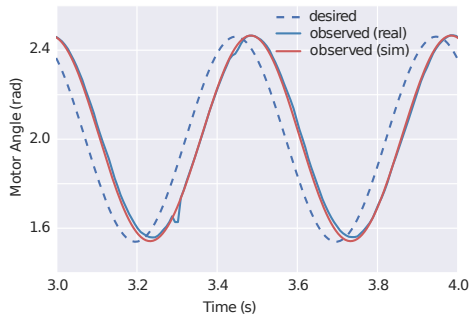}
    \caption{Comparison of the simulated motor trajectory (red) with the ground truth (blue).}
    \label{fig:motor_test}
\end{figure}

We performed an experiment to validate the new actuator model. We actuated a motor with a desired trajectory of sine curve. With this actuator model, the simulated trajectory agrees with the ground truth (Figure \ref{fig:motor_test}).

\paragraph{Latency} Latency is one of the main causes of instability for feedback control. It is the time delay between when a motor command is sent that causes the state of the robot to change and when the sensor measurement of this change is reported back to the controller. In Bullet, the motor commands take effect immediately and the sensors report back the state instantaneously. This instantaneous feedback makes the stability region of a feedback controller in simulation much larger than its implementation on hardware. For this reason, we often see a feedback policy learned in simulation starts to oscillate, diverge and ultimately fail in the real world.

To model latency, we keep a history of observations and their measurement time $\{(t_i, \mathbf{O}_i)_{i=0,1,...,n-1}\}$, where $t_i = i\Delta t$ and $\Delta t$ is the time step. At the current step $n$, when the controller needs an observation, we search the history for two adjacent observations $\mathbf{O}_i$ and $\mathbf{O}_{i+1}$ where $t_i \leq n\Delta t-t_{\textrm{latency}} \leq t_{i+1}$ and linearly interpolate them. 

To measure the latency on the physical system, we send a spike of PWM signal that lasts for one time step, which causes a small movement of the motor. We measure the time delay between when the spike is sent and when the resultant motor movement is reported. Note that we have two different latencies for the microcontroller and Nvidia Jetson TX2. The PD servo running on the microcontroller has a lower latency (3ms) while the locomotion controller executed on TX2 has a higher latency (typically 15-19ms). We use these measurements to set the correct latencies in simulation.

\subsection{Learning Robust Controllers}
\label{sec:robust}
Robust control aims at achieving robust performance in the presence of model error. Thus, it is easier to transfer a robust controller to the real world even if the simulated dynamics are not identical to their real-world counterparts. In this paper, we experiment with three different ways to learn a robust controller: randomizing dynamic parameters, adding random perturbations and using a compact observation space.

Prior studies showed that randomizing the dynamic parameters during training can increase the robustness of the learned controller \cite{peng2017sim,rajeswaran2016epopt}. At the beginning of each training episode, we randomly sample a set of physical parameters and use them in the simulation. The samples are drawn uniformly within certain ranges. The complete list of randomized parameters and their ranges of randomization is summarized in Table \ref{table:randomization}. 

Since dynamics randomization trades optimality for robustness \cite{luo2017robust}, we choose the parameters and their ranges in Table \ref{table:randomization} carefully to prevent learning overly conservative running gaits. Mass and motor friction are commonly randomized parameters \cite{rajeswaran2016epopt,peng2017sim}. While we give them conservative ranges since we have measured them during system identification, we are less certain about inertia because it is estimated based on a uniform density assumption. Also some quantities can change over time. For example, the motor strength can vary due to wear and tear. The control step and latency can fluctuate because of the non real-time system. The battery voltage can change based on whether it is fully charged. For the reasons stated above, we choose to randomize these parameters and their ranges based on the actual measurements plus small safety margins. Performing accurate identification for contact parameters is challenging. While the Bullet physics library uses an LCP-based contact model and provides a number of tunable parameters, we choose to focus on the lateral friction and leave others in their default values. We randomly sample the friction coefficient between 0.5 and 1.25 because this is the typical range of friction coefficient between the Minitaur's rubber feet and various carpet floors \cite{Slipperiness}. Since the real IMU measurement often carries bias and noise, we also add a small amount of Gaussian noise to the simulated IMU readings. 

\begin{table}
\centering
\caption{Randomized physical parameters and their ranges.}
  \begin{tabular}{| l | c | r |}
  \hline
  parameter & lower bound & upper bound \\
    \hline
    mass & 80\% & 120\% \\ 
    motor friction & 0Nm & 0.05Nm \\
    inertia & 50\% & 150\% \\
    motor strength & 80\% & 120\% \\
    control step & 3ms & 20ms \\
    latency & 0ms & 40ms \\
    battery voltage & 14.0V & 16.8V \\
    contact friction & 0.5 & 1.25 \\
    IMU bias & -0.05 radian & 0.05 radian \\
    IMU noise (std) & 0 radian & 0.05 radian \\
    \hline
  \end{tabular}
  \label{table:randomization}
  
\end{table}

Another popular method to improve the robustness of a learned controller is to add random perturbations \cite{pinto2017robust}. During training, we add a perturbation force to the base of the simulated Minitaur every 200 steps of simulation (1.2s). The perturbation force lasts for ten steps (0.06s), has a random direction and a random magnitude ranging from 130N to 220N. The perturbations can knock the simulated Minitaur out of balance so that it needs to learn how to recover balance from different situations. 

We find that the design of the observation space also plays an important role to narrow the reality gap. If the observation space is high dimensional, the learned policy can easily overfit the simulated environment, which makes it difficult to transfer to the real robots. In this paper, we use a compact observation space (Section \ref{sec:obs}) that leaves less room for overfitting to unimportant details of the simulation. A detailed analysis on the choice of observation space is presented in Section \ref{sec:gap_analysis}.

\begin{figure}
    \centering
    \includegraphics[width=2.3in]{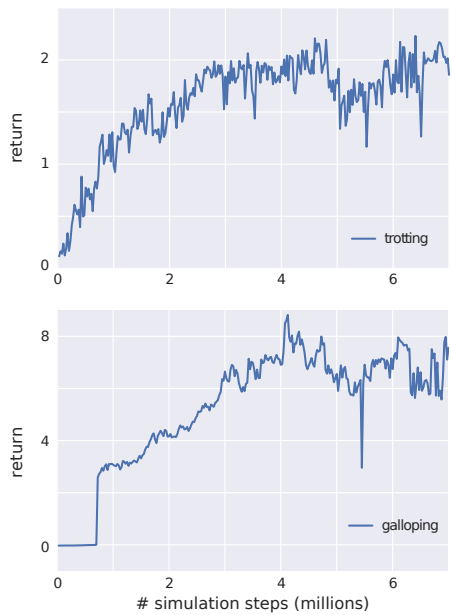}
    \caption{The learning curves of trotting and galloping.}
    \label{fig:learning_curves}
\end{figure}

\begin{table}[!t]
\centering
\caption{Parameters of learning algorithm for each task.}
  \begin{tabular}{| c | c | c | c | c |}
  \hline
  Gait & Observation & Policy Net & Value Net & Learning\\
   & Dimension & Size & Size & Time (Hour) \\
    \hline
    Trotting  & 4 & (125, 89) & (89, 55) & 4.35\\
    Galloping & 12 & (185, 95) & (95, 85) & 3.25\\
    \hline
  \end{tabular}
  \label{table:learningParameters}
\end{table}

\section{Evaluation and Discussion}
\label{sec:evaluation}
We tested our system with two locomotion tasks: galloping and trotting. Please watch the accompanying video\footnote{https://www.youtube.com/watch?v=lUZUr7jxoqM} for the learned locomotion gaits. We first learned the locomotion controllers in simulation. The simulated Minitaur environments are open-sourced in Bullet physics library\footnote{https://git.io/vp0V3}. We represented both the policy and the value functions as fully connected neural networks, each of which had two hidden layers. Their sizes were determined using hyperparameter search. At each iteration of policy update, we collected the simulated experience by running 25 roll-outs (up to 1000 steps each) in parallel. The training terminated after the maximum number of simulation steps (7 million) had been reached. The learning curves of trotting and galloping are shown in Figure \ref{fig:learning_curves}, and the parameter settings are summarized in Table \ref{table:learningParameters}. After the policies were learned, we deployed them on the real robot. The controllers worked directly in the real world without additional fine tuning on the physical system.

\subsection{Locomotion Tasks}

In the first experiment, we let the system learn from scratch: We set the open loop component $\bar{\mathbf{a}}(t)=0$ and gave the feedback component large output bounds: The bounds for swing are $[-0.5, 0.5]$ radians and the bounds for extension are $[\frac{\pi}{2}-0.5, \frac{\pi}{2}+0.5]$ radians. Initially, with the baseline simulation (no actuator model and latency), our system did not come up with an agile gait. Instead, a slow walking gait was learned in simulation. This might be caused by the default constraint-based actuator model in Bullet, which seems to be overdamped and limits the agility of the motion. Moreover, the real Minitaur fell to the ground immediately due to the reality gap. After we improved the simulation (Section \ref{sec:improve_sim}), an agile galloping gait emerged automatically. Galloping is the fastest gait for quadrupeds in nature. The running speed of the Minitaur reaches roughly 1.34 m/s (2.48 body lengths per second) in simulation and 1.18 m/s (2.18 body lengths per second) in the real world. We repeated the training with different hyperparameters and random seeds, and found that the majority of the solutions converged to galloping. We also observed a small number of other gaits, including trotting, pacing, and gaits that are uncommon among quadruped animals.

In the second experiment, we would like the Minitaur to learn trotting, an intermediate-speed running gait commonly adopted by horses (and many other quadrupeds) in which the diagonally opposite legs move together. While it is unclear how to use reward shaping to learn such a gait, we can directly control the learned gait by providing an open loop signal ($\bar{\mathbf{a}}(t)$ in eq. (\ref{eq:policy})). The signal consists of two sine curves.
\begin{eqnarray}
\nonumber
\bar{s}(t) & = & 0.3 \sin(4\pi t) \\
\nonumber
\bar{e}(t) & = & 0.35 \sin(4\pi t) + 2
\end{eqnarray}
where $\bar{s}(t)$ is the signal for the swing and the $\bar{e}(t)$ is for the extension. One diagonal pair of legs shares the same curves and the other diagonal pair's are 180 degree out of phase. Note that while this open loop controller expresses the user's preference of the locomotion style, by itself, it cannot produce any forward movement in the real world: The Minitaur loses balance immediately and sits down on its rear legs. To maintain balance and keep the learned gait similar to the user-specified signal, we used small output bounds of the feedback controller: $[-0.25, 0.25]$ radians for both the swing and the extension degrees of freedom. After training with the improved simulator and random perturbations, the Minitaur is able to trot stably in simulation. However, when the policies were deployed on the robot, we had mixed results due to the reality gap: Some policies can transfer while others cannot. We further reduced the observation space to four dimensional, keeping only the IMU readings (the roll, pitch and the angular velocities of the base along these two axes) and retrained in simulation. This time, we observed stable, comparable movements in both simulation and on the real robot. The trotting speed is roughly 0.50 m/s (0.93 body lengths per second) in simulation and 0.60 m/s (1.11 body lengths per second) in the real world. 

\begin{table}[!b]
\centering
\caption{Comparisons between learned and handcrafted gaits.}
  \begin{tabular}{| l | c | r |}
  \hline
  Gait & Speed (m/s) & Avg. Mech. Power (watt) \\
    \hline
    Trotting (handcrafted) & 0.56 & 92.72 \\
    Trotting (learned) & \textbf{0.60} &  \textbf{71.78} \\
    \hline
    Galloping (handcrafted) & \textbf{1.21} & 290.00 \\
    Galloping (learned) & 1.18 & \textbf{188.79} \\
    \hline
  \end{tabular}
  \label{table:gaitsComparison}
\end{table}

We compared the learned gaits with the handcrafted ones from Ghost Robotics \cite{de2017}. Table~\ref{table:gaitsComparison} summarizes the speed and the energy consumption of these gaits on the real robots. While our learned gaits are as fast as the ones carefully tuned by experts, they consume significantly less power ($35\%$ and $23\%$ reduction for galloping and trotting respectively).

\subsection{Narrowing the Reality Gap}
\label{sec:gap_analysis}

As discussed in Section \ref{sec:robust}, we narrow the reality gap by improving the simulation fidelity and learning robust controllers. Using trotting as a test case, we performed comprehensive evaluations to understand how each of these techniques, improving simulation, adding randomization and using compact observation space, can narrow the reality gap. 

First, we defined a quantitative measure of the reality gap. One common choice is the success rate \cite{mordatch2015ensemble}: The percentage of the controllers that can balance in the real world for the entire episode (1000 steps, about 6 seconds). However, the binary outcome of success or failure does not capture the key characteristics of locomotion, such as running speed and energy consumption. For this reason, we adopt a continuous measure used by Koos et al. \cite{koos2010crossing}. We compute the expected return from eq. (\ref{eq:reward}), both in simulation and in the real experiments. Their difference measures the reality gap. 

We observed a large variance of the controller performance in the real experiments due to the randomness of the learning algorithm, the stochastic nature of the real world, and the differences in robustness of the learned controllers. To increase the confidence of our conclusions, we performed a large number of learning trials and robot experiments before a conclusion is reached. In each of the experiments below, we trained 100 controllers with different hyperparameters and random seeds. We deployed the top three controllers on the Minitaur based on their returns in simulation. Each controller was run three times. The average return of these nine runs was reported as the expected return.

\begin{figure}[!t]
    \centering
    \includegraphics[width=2.2in]{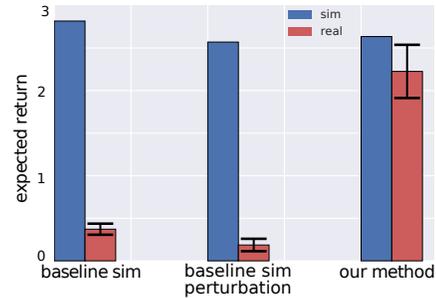}
    \caption{Controller performance in simulation (blue) and on the robot (red). From left to right, the controllers are trained using baseline simulation, using baseline simulation with random perturbations, and using improved simulation with random perturbations. Error bars indicate one standard error.}
    \label{fig:improve_simulation}
\end{figure}

To examine how improving the simulation can narrow the reality gap, we trained three groups of controllers, all using the four-dimensional observations space. In the first group, the controllers were trained using the baseline simulation (no actuator model and latency handling). The second group used the baseline simulation with random perturbations during training. The last group was our method, which used the improved simulation and random perturbations. The top three controllers in each group all performed well in simulation. However, when they were deployed on the robot, the controllers in group I and II performed poorly. From the left and the center bars of Figure \ref{fig:improve_simulation}, we can clearly see a large reality gap (the difference between the blue and the red bars). In contrast, the controllers in group III performed comparably in simulation and in the real world (the right bars in Figure \ref{fig:improve_simulation}). This indicates that the improvement of the simulation is essential to our system. If the model discrepancy is too big, even a robust controller trained with random perturbations cannot overcome the reality gap. We also found that both accurate actuator model and latency simulation are important. Without either of them, the learned controllers do not work on the real robot.

\begin{figure}[!t]
    \centering
    \includegraphics[width=2.6in]{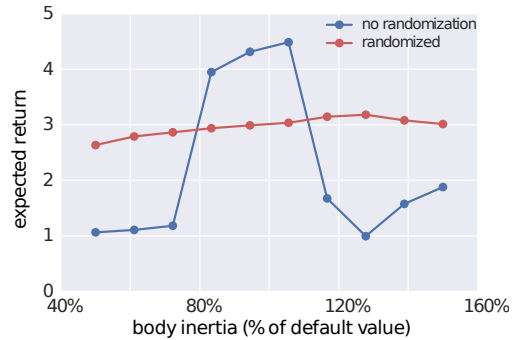}
    \caption{Performance comparison of controllers that are trained with (red) and without (blue) randomization and tested with different body inertia. }
    \label{fig:sensitivity_inertia}
\end{figure}

Next, we evaluated the impact of randomizing physical parameters and adding random perturbations on the robustness and transferability of controllers. Experiments showed that these two approaches had similar effects. In fact, as pointed out in the prior work \cite{pinto2017robust}, uncertainties of the physical parameters can be viewed as extra forces/torques applied to the system. For this reason, we grouped their results and reported them under the common name of ``randomization''. We first evaluated the trade-off between robustness and optimality using randomization. Given a trained controller, we computed its expected return in a set of different test environments in simulation: For each parameter in Table \ref{table:randomization}, we evenly sampled 10 values within its range while keeping other parameters unchanged. We used these samples to construct different test environments. We computed the return of the controller in each test environment and observed how the performance changed across environments. For example, we tested two controllers, trained with or without randomization, in the environments that the inertia of the robot ranges from $50\%$ to $150\%$ of its default value. Figure \ref{fig:sensitivity_inertia} shows that the controller trained with randomization (red) performs similarly across all these environments. In contrast, the performance of the controller trained without randomization can drop significantly if the actual inertia is different from what it was trained for. However, it has a higher peak performance.

\begin{figure}[!t]
    \centering
    \includegraphics[width=2.9in]{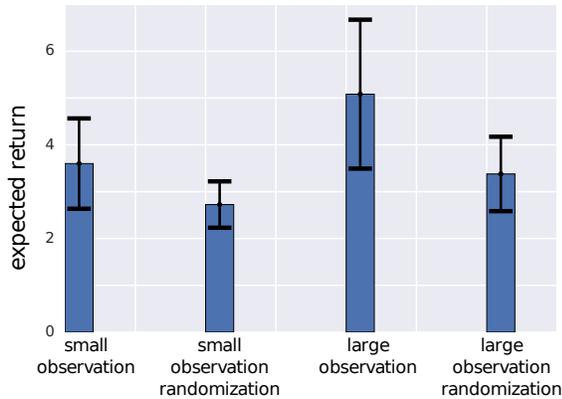}
    \caption{Performance of controllers when they are tested in different simulation environments. Error bars indicate one standard deviation.}
    \label{fig:sensitivity_all}
\end{figure}

\begin{figure}[!t]
    \centering
    \includegraphics[width=2.9in]{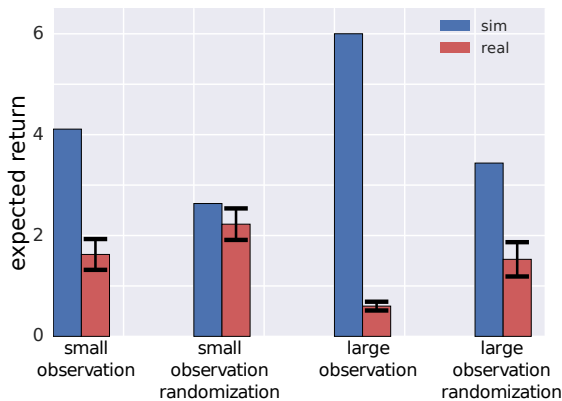}
    \caption{Comparison of controllers trained with different observation spaces and randomization. The blue and red bars are the performance in simulation and in the real world respectively. Error bars indicate one standard error.}
    \label{fig:observation_randomization}
\end{figure}

We aggregated these results and calculated the mean and the standard deviation of the returns across all the test environments. It is obvious in Figure \ref{fig:sensitivity_all} that regardless of the choice of observation space, the returns of the controllers trained using randomization have a lower mean and a lower standard deviation (Refer to the first two bars in Figure \ref{fig:sensitivity_all} for the case of small observation space, and last two bars for the case of large observation space). Lower mean indicates suboptimal performance. The controllers learn to behave conservatively in the stochastic environments. Lower standard deviation indicates robustness. The performance varies less even when the controllers are tested in different environments. These results demonstrate a clear trade-off between robustness and optimality when randomization is used. Since randomization is not a free meal, we should use it only when necessary.

We then evaluated the impact of randomization on the transferability of controllers. Figure \ref{fig:observation_randomization} shows that regardless of the choice of observations, randomization helps to narrow the reality gap. When combining it with the small observation space, we achieved the best results in the real world: All nine runs of the top three controllers in this group can trot more than three meters and balance for the entire episode.

Lastly, we examined the relation between the reality gap and the choice of observations. In this experiment, we trained controllers with two different observation space. The small space (4D) only consists of the IMU information: the roll, pitch, and the angular velocities of the base along these two axes. The large space (12D) consists of both the IMU information and the motor angles. We found that in simulation, when the observation space is large, the performance of the learned controllers is higher (Compare the blue bars between 1st and 3rd, and between 2nd and 4th in Figure \ref{fig:observation_randomization}). We suspect that the tasks become easier to learn because the policy can take advantage of the more information in the large observation space. However, when these policies are deployed on the robot, the result is the opposite. Controllers with the large observation space perform worse in the real world and the reality gap is wider (compare the red bars between 1st and 3rd, and between 2nd and 4th in Figure \ref{fig:observation_randomization}). We believe that this is caused by the mismatch of the observation distributions between training in simulation and testing on the physical system. When the space is large, the observations encountered in training become relatively sparse. Encountering similar observations in the real world is less likely, which can cause the robot to fall. In contrast, when the observation space is small, the observation distributions in training and testing are more likely to overlap, which narrows the reality gap.

\section{Conclusion} 
We have shown that deep RL can be applied to learn agile locomotion automatically for robots. We have presented a complete system that applies deep RL in simulated environments, which can learn from scratch or can allow users to guide the learning process. With an accurate physical model and robust controllers, we have successfully deployed the controllers learned in simulation on the real robots. Using this system, we are able to develop two agile locomotion gaits, trotting and galloping, for a quadruped robot.

The focus of our paper is on learning transferable locomotion policies. For this purpose, we used a simple reward function and a simple environment: maximizing the running speed on a flat ground. In real-world scenarios, robots need to see the environment, adjust its speed and turn agilely to navigate the complex physical world. This points us to two interesting avenues for future work. First, we would like to learn locomotion policies that can dynamically change running speed and direction. Second, it would be interesting to extend this work to handle complex terrain structures by integrating vision as part of sensory input.

%% Use plainnat to work nicely with natbib. 

%\bibliographystyle{plainnat}
\bibliographystyle{unsrtnat}
\bibliography{references}

\begin{thebibliography}{56}
\providecommand{\natexlab}[1]{#1}
\providecommand{\url}[1]{\texttt{#1}}
\expandafter\ifx\csname urlstyle\endcsname\relax
  \providecommand{\doi}[1]{doi: #1}\else
  \providecommand{\doi}{doi: \begingroup \urlstyle{rm}\Url}\fi

\bibitem[Raibert(1986)]{Raibert:1986:LRB:6152}
Marc~H. Raibert.
\newblock \emph{Legged Robots That Balance}.
\newblock Massachusetts Institute of Technology, Cambridge, MA, USA, 1986.
\newblock ISBN 0-262-18117-7.

\bibitem[Pratt and Pratt(1998)]{pratt1998intuitive}
Jerry Pratt and Gill Pratt.
\newblock Intuitive control of a planar bipedal walking robot.
\newblock In \emph{Robotics and Automation, 1998. Proceedings. 1998 IEEE
  International Conference on}, volume~3, pages 2014--2021. IEEE, 1998.

\bibitem[De(2017)]{de2017}
Avik De.
\newblock Modular hopping and running via parallel composition, 2017.

\bibitem[Lillicrap et~al.(2015)Lillicrap, Hunt, Pritzel, Heess, Erez, Tassa,
  Silver, and Wierstra]{lillicrap2015continuous}
Timothy~P Lillicrap, Jonathan~J Hunt, Alexander Pritzel, Nicolas Heess, Tom
  Erez, Yuval Tassa, David Silver, and Daan Wierstra.
\newblock Continuous control with deep reinforcement learning.
\newblock \emph{arXiv preprint arXiv:1509.02971}, 2015.

\bibitem[Schulman et~al.(2017)Schulman, Wolski, Dhariwal, Radford, and
  Klimov]{DBLP:journals/corr/SchulmanWDRK17}
John Schulman, Filip Wolski, Prafulla Dhariwal, Alec Radford, and Oleg Klimov.
\newblock Proximal policy optimization algorithms.
\newblock \emph{CoRR}, abs/1707.06347, 2017.

\bibitem[Duan et~al.(2016)Duan, Chen, Houthooft, Schulman, and
  Abbeel]{Duan:2016:BDR:3045390.3045531}
Yan Duan, Xi~Chen, Rein Houthooft, John Schulman, and Pieter Abbeel.
\newblock Benchmarking deep reinforcement learning for continuous control.
\newblock In \emph{Proceedings of the 33rd International Conference on
  International Conference on Machine Learning - Volume 48}, ICML'16, pages
  1329--1338. JMLR.org, 2016.

\bibitem[Koos et~al.(2010)Koos, Mouret, and Doncieux]{koos2010crossing}
Sylvain Koos, Jean-Baptiste Mouret, and St{\'e}phane Doncieux.
\newblock Crossing the reality gap in evolutionary robotics by promoting
  transferable controllers.
\newblock In \emph{Proceedings of the 12th annual conference on Genetic and
  evolutionary computation}, pages 119--126. ACM, 2010.

\bibitem[Boeing and Br{\"a}unl(2012)]{boeing2012leveraging}
Adrian Boeing and Thomas Br{\"a}unl.
\newblock Leveraging multiple simulators for crossing the reality gap.
\newblock In \emph{Control Automation Robotics \& Vision (ICARCV), 2012 12th
  International Conference on}, pages 1113--1119. IEEE, 2012.

\bibitem[Levine et~al.()Levine, Pastor, Krizhevsky, Ibarz, and
  Quillen]{levine2016learning}
Sergey Levine, Peter Pastor, Alex Krizhevsky, Julian Ibarz, and Deirdre
  Quillen.
\newblock Learning hand-eye coordination for robotic grasping with deep
  learning and large-scale data collection.
\newblock \emph{The International Journal of Robotics Research}.

\bibitem[Silver et~al.(2017)Silver, Schrittwieser, Simonyan, Antonoglou, Huang,
  Guez, Hubert, Baker, Lai, Bolton, et~al.]{silver2017mastering}
David Silver, Julian Schrittwieser, Karen Simonyan, Ioannis Antonoglou, Aja
  Huang, Arthur Guez, Thomas Hubert, Lucas Baker, Matthew Lai, Adrian Bolton,
  et~al.
\newblock Mastering the game of {G}o without human knowledge.
\newblock \emph{Nature}, 550\penalty0 (7676):\penalty0 354, 2017.

\bibitem[Kober and Peters(2012)]{kober2012reinforcement}
Jens Kober and Jan Peters.
\newblock Reinforcement learning in robotics: A survey.
\newblock In \emph{Reinforcement Learning}, pages 579--610. Springer, 2012.

\bibitem[Choromanski et~al.(2018)Choromanski, Iscen, Sindhwani, Tan, and
  Coumans]{icra18-minitaur}
Krzysztof Choromanski, Atil Iscen, Vikas Sindhwani, Jie Tan, and Erwin Coumans.
\newblock Optimizing simulations with noise-tolerant structured exploration.
\newblock In \emph{Robotics and Automation (ICRA), 2018 IEEE International
  Conference on}. IEEE, 2018.

\bibitem[Cully et~al.(2015)Cully, Clune, Tarapore, and Mouret]{cully2015robots}
Antoine Cully, Jeff Clune, Danesh Tarapore, and Jean-Baptiste Mouret.
\newblock Robots that can adapt like animals.
\newblock \emph{Nature}, 521\penalty0 (7553):\penalty0 503, 2015.

\bibitem[Calandra et~al.(2016)Calandra, Seyfarth, Peters, and
  Deisenroth]{calandra2016bayesian}
Roberto Calandra, Andr{\'e} Seyfarth, Jan Peters, and Marc~Peter Deisenroth.
\newblock Bayesian optimization for learning gaits under uncertainty.
\newblock \emph{Annals of Mathematics and Artificial Intelligence}, 76\penalty0
  (1-2):\penalty0 5--23, 2016.

\bibitem[Antonova et~al.(2017)Antonova, Rai, and
  Atkeson]{DBLP:conf/corl/AntonovaRA17}
Rika Antonova, Akshara Rai, and Christopher~G. Atkeson.
\newblock Deep kernels for optimizing locomotion controllers.
\newblock In \emph{1st Annual Conference on Robot Learning, CoRL}, pages
  47--56, 2017.

\bibitem[Brockman et~al.(2016)Brockman, Cheung, Pettersson, Schneider,
  Schulman, Tang, and Zaremba]{1606.01540}
Greg Brockman, Vicki Cheung, Ludwig Pettersson, Jonas Schneider, John Schulman,
  Jie Tang, and Wojciech Zaremba.
\newblock Open{AI} {G}ym, 2016.

\bibitem[Tassa et~al.(2018)Tassa, Doron, Muldal, Erez, Li, Casas, Budden,
  Abdolmaleki, Merel, Lefrancq, et~al.]{tassa2018deepmind}
Yuval Tassa, Yotam Doron, Alistair Muldal, Tom Erez, Yazhe Li, Diego de~Las
  Casas, David Budden, Abbas Abdolmaleki, Josh Merel, Andrew Lefrancq, et~al.
\newblock Deep{M}ind control suite.
\newblock \emph{arXiv preprint arXiv:1801.00690}, 2018.

\bibitem[Hafner et~al.(2017)Hafner, Davidson, and Vanhoucke]{hafner2017agents}
Danijar Hafner, James Davidson, and Vincent Vanhoucke.
\newblock {TensorFlow} agents: Efficient batched reinforcement learning in
  {TensorFlow}.
\newblock \emph{arXiv preprint arXiv:1709.02878}, 2017.

\bibitem[Benbrahim and Franklin(1997)]{benbrahimbiped}
Hamid Benbrahim and Judy~A Franklin.
\newblock Biped dynamic walking using reinforcement learning.
\newblock \emph{Robotics and Autonomous Systems}, 22\penalty0 (3-4):\penalty0
  283--302, 1997.

\bibitem[Tedrake et~al.()Tedrake, Zhang, and Seung]{tedrake2004stochastic}
Russ Tedrake, Teresa~Weirui Zhang, and H~Sebastian Seung.
\newblock Stochastic policy gradient reinforcement learning on a simple 3d
  biped.
\newblock In \emph{Intelligent Robots and Systems, 2004.(IROS 2004).
  Proceedings. 2004 IEEE/RSJ International Conference on}, volume~3, pages
  2849--2854. IEEE.

\bibitem[Endo et~al.(2005)Endo, Morimoto, Matsubara, Nakanishi, and
  Cheng]{endo2005learning}
Gen Endo, Jun Morimoto, Takamitsu Matsubara, Jun Nakanishi, and Gordon Cheng.
\newblock Learning {CPG} sensory feedback with policy gradient for biped
  locomotion for a full-body humanoid.
\newblock In \emph{Proceedings of the 20th national conference on Artificial
  intelligence-Volume 3}, pages 1267--1273. AAAI Press, 2005.

\bibitem[Kohl and Stone(2004)]{icra04}
Nate Kohl and Peter Stone.
\newblock Policy gradient reinforcement learning for fast quadrupedal
  locomotion.
\newblock In \emph{Proceedings of the {IEEE} International Conference on
  Robotics and Automation}, 2004.

\bibitem[Ogino et~al.(2004)Ogino, Katoh, Aono, Asada, and
  Hosoda]{DBLP:journals/ar/OginoKAAH04}
Masaki Ogino, Yutaka Katoh, Masahiro Aono, Minoru Asada, and Koh Hosoda.
\newblock Reinforcement learning of humanoid rhythmic walking parameters based
  on visual information.
\newblock \emph{Advanced Robotics}, 18\penalty0 (7):\penalty0 677--697, 2004.

\bibitem[Gay et~al.(2013)Gay, Santos-Victor, and Ijspeert]{gay2013learning}
S{\'e}bastien Gay, Jos{\'e} Santos-Victor, and Auke Ijspeert.
\newblock Learning robot gait stability using neural networks as sensory
  feedback function for central pattern generators.
\newblock In \emph{Intelligent Robots and Systems (IROS), 2013 IEEE/RSJ
  International Conference on}, pages 194--201. IEEE, 2013.

\bibitem[Levine and Koltun(2014)]{2014-cgps}
Sergey Levine and Vladlen Koltun.
\newblock Learning complex neural network policies with trajectory
  optimization.
\newblock In \emph{ICML '14: Proceedings of the 31st International Conference
  on Machine Learning}, 2014.

\bibitem[Peng et~al.(2015)Peng, Berseth, and van~de Panne]{2015-TOG-terrainRL}
Xue~Bin Peng, Glen Berseth, and Michiel van~de Panne.
\newblock Dynamic terrain traversal skills using reinforcement learning.
\newblock \emph{ACM Trans. Graph.}, 34\penalty0 (4):\penalty0 80:1--80:11,
  2015.
\newblock ISSN 0730-0301.
\newblock \doi{10.1145/2766910}.

\bibitem[Peng et~al.(2016)Peng, Berseth, and van~de Panne]{2016-TOG-deepRL}
Xue~Bin Peng, Glen Berseth, and Michiel van~de Panne.
\newblock Terrain-adaptive locomotion skills using deep reinforcement learning.
\newblock \emph{ACM Transactions on Graphics (Proc. SIGGRAPH 2016)},
  35\penalty0 (4), 2016.

\bibitem[Peng et~al.(2017{\natexlab{a}})Peng, Berseth, Yin, and van~de
  Panne]{2017-TOG-deepLoco}
Xue~Bin Peng, Glen Berseth, KangKang Yin, and Michiel van~de Panne.
\newblock {DeepLoco}: Dynamic locomotion skills using hierarchical deep
  reinforcement learning.
\newblock \emph{ACM Transactions on Graphics (Proc. SIGGRAPH 2017)},
  36\penalty0 (4), 2017{\natexlab{a}}.

\bibitem[Heess et~al.(2017)Heess, TB, Sriram, Lemmon, Merel, Wayne, Tassa,
  Erez, Wang, Eslami, Riedmiller, and
  Silver]{DBLP:journals/corr/HeessTSLMWTEWER17}
Nicolas Heess, Dhruva TB, Srinivasan Sriram, Jay Lemmon, Josh Merel, Greg
  Wayne, Yuval Tassa, Tom Erez, Ziyu Wang, S.~M.~Ali Eslami, Martin~A.
  Riedmiller, and David Silver.
\newblock Emergence of locomotion behaviours in rich environments.
\newblock \emph{CoRR}, abs/1707.02286, 2017.

\bibitem[Sharma and Kitani(2018)]{Sharma-2018-103569}
Arjun Sharma and Kris~M. Kitani.
\newblock Phase-parametric policies for reinforcement learning in cyclic
  environments.
\newblock In \emph{AAAI Conference on Artificial Intelligence}, Pittsburgh, PA,
  2018.

\bibitem[Neunert et~al.(2017)Neunert, Boaventura, and Buchli]{neunert2017off}
Michael Neunert, Thiago Boaventura, and Jonas Buchli.
\newblock Why off-the-shelf physics simulators fail in evaluating feedback
  controller performance-a case study for quadrupedal robots.
\newblock In \emph{Advances in Cooperative Robotics}, pages 464--472. World
  Scientific, 2017.

\bibitem[Zhu et~al.(2017)Zhu, Kimmel, Bekris, and
  Boularias]{DBLP:journals/corr/abs-1710-08893}
Shaojun Zhu, Andrew Kimmel, Kostas~E. Bekris, and Abdeslam Boularias.
\newblock Model identification via physics engines for improved policy search.
\newblock \emph{CoRR}, abs/1710.08893, 2017.

\bibitem[Li et~al.(2013)Li, Zhang, and Goldman]{li2013terradynamics}
Chen Li, Tingnan Zhang, and Daniel~I Goldman.
\newblock A terradynamics of legged locomotion on granular media.
\newblock \emph{Science}, 339\penalty0 (6126):\penalty0 1408--1412, 2013.

\bibitem[Tan et~al.(2016)Tan, Xie, Boots, and Liu]{tan2016simulation}
Jie Tan, Zhaoming Xie, Byron Boots, and C.~Karen Liu.
\newblock Simulation-based design of dynamic controllers for humanoid
  balancing.
\newblock In \emph{Intelligent Robots and Systems (IROS), 2016 IEEE/RSJ
  International Conference on}, pages 2729--2736. IEEE, 2016.

\bibitem[M{\"o}ckel et~al.(2013)M{\"o}ckel, Yura, The~Nguyen, Vespignani,
  Bonardi, Pouya, Spr{\"o}witz, van~den Kieboom, Wilhelm, and
  Ijspeert]{escidoc:2316379}
Rico M{\"o}ckel, N.~Perov Yura, Anh The~Nguyen, Massimo Vespignani, Stephane
  Bonardi, Soha Pouya, Alexander Spr{\"o}witz, Jesse van~den Kieboom, Frederic
  Wilhelm, and Auke~Jan Ijspeert.
\newblock Gait optimization for roombots modular robots - matching simulation
  and reality.
\newblock In \emph{Proceedings of the 2013 IEEE/RSJ International Conference on
  Intelligent Robots and Systems}, pages 3265--3272, Tokyo, 2013. IEEE.

\bibitem[Bongard et~al.(2006)Bongard, Zykov, and Lipson]{Bongard2006}
Josh Bongard, Victor Zykov, and Hod Lipson.
\newblock Resilient machines through continuous self-modeling.
\newblock \emph{Science}, 314:\penalty0 1118--21, 2006.

\bibitem[Ha and Yamane(2015)]{ha2015reducing}
Sehoon Ha and Katsu Yamane.
\newblock Reducing hardware experiments for model learning and policy
  optimization.
\newblock In \emph{Robotics and Automation (ICRA), 2015 IEEE International
  Conference on}, pages 2620--2626. IEEE, 2015.

\bibitem[Yu et~al.(2017)Yu, Tan, Liu, and Turk]{DBLP:journals/corr/YuLT17}
Wenhao Yu, Jie Tan, C.~Karen Liu, and Greg Turk.
\newblock Preparing for the unknown: Learning a universal policy with online
  system identification.
\newblock \emph{CoRR}, abs/1702.02453, 2017.

\bibitem[Peng et~al.(2017{\natexlab{b}})Peng, Andrychowicz, Zaremba, and
  Abbeel]{peng2017sim}
Xue~Bin Peng, Marcin Andrychowicz, Wojciech Zaremba, and Pieter Abbeel.
\newblock Sim-to-real transfer of robotic control with dynamics randomization.
\newblock \emph{arXiv preprint arXiv:1710.06537}, 2017{\natexlab{b}}.

\bibitem[Jakobi et~al.(1995)Jakobi, Husbands, and Harvey]{jakobi1995noise}
Nick Jakobi, Phil Husbands, and Inman Harvey.
\newblock Noise and the reality gap: The use of simulation in evolutionary
  robotics.
\newblock \emph{Advances in artificial life}, pages 704--720, 1995.

\bibitem[Pinto et~al.(2017)Pinto, Davidson, Sukthankar, and
  Gupta]{pinto2017robust}
Lerrel Pinto, James Davidson, Rahul Sukthankar, and Abhinav Gupta.
\newblock Robust adversarial reinforcement learning.
\newblock \emph{arXiv preprint arXiv:1703.02702}, 2017.

\bibitem[Tobin et~al.(2017)Tobin, Fong, Ray, Schneider, Zaremba, and
  Abbeel]{tobin2017domain}
Josh Tobin, Rachel Fong, Alex Ray, Jonas Schneider, Wojciech Zaremba, and
  Pieter Abbeel.
\newblock Domain randomization for transferring deep neural networks from
  simulation to the real world.
\newblock \emph{arXiv preprint arXiv:1703.06907}, 2017.

\bibitem[Mordatch et~al.(2015)Mordatch, Lowrey, and
  Todorov]{mordatch2015ensemble}
Igor Mordatch, Kendall Lowrey, and Emanuel Todorov.
\newblock Ensemble-{CIO}: Full-body dynamic motion planning that transfers to
  physical humanoids.
\newblock In \emph{Intelligent Robots and Systems (IROS), 2015 IEEE/RSJ
  International Conference on}, pages 5307--5314. IEEE, 2015.

\bibitem[Rajeswaran et~al.(2016)Rajeswaran, Ghotra, Ravindran, and
  Levine]{rajeswaran2016epopt}
Aravind Rajeswaran, Sarvjeet Ghotra, Balaraman Ravindran, and Sergey Levine.
\newblock {EPO}pt: Learning robust neural network policies using model
  ensembles.
\newblock \emph{arXiv preprint arXiv:1610.01283}, 2016.

\bibitem[Rusu et~al.(2016)Rusu, Vecerik, Roth{\"{o}}rl, Heess, Pascanu, and
  Hadsell]{DBLP:journals/corr/RusuVRHPH16}
Andrei~A. Rusu, Matej Vecerik, Thomas Roth{\"{o}}rl, Nicolas Heess, Razvan
  Pascanu, and Raia Hadsell.
\newblock Sim-to-real robot learning from pixels with progressive nets.
\newblock \emph{CoRR}, abs/1610.04286, 2016.

\bibitem[Hanna and Stone(2017)]{AAAI17-Hanna}
Josiah Hanna and Peter Stone.
\newblock Grounded action transformation for robot learning in simulation.
\newblock In \emph{Proceedings of the 31st AAAI Conference on Artificial
  Intelligence (AAAI)}, 2017.

\bibitem[Tzeng et~al.(2015)Tzeng, Devin, Hoffman, Finn, Peng, Levine, Saenko,
  and Darrell]{DBLP:journals/corr/TzengDHFPLSD15}
Eric Tzeng, Coline Devin, Judy Hoffman, Chelsea Finn, Xingchao Peng, Sergey
  Levine, Kate Saenko, and Trevor Darrell.
\newblock Towards adapting deep visuomotor representations from simulated to
  real environments.
\newblock \emph{CoRR}, abs/1511.07111, 2015.

\bibitem[Fang et~al.(2017)Fang, Bai, Hinterstoisser, and
  Kalakrishnan]{DBLP:journals/corr/abs-1710-06422}
Kuan Fang, Yunfei Bai, Stefan Hinterstoisser, and Mrinal Kalakrishnan.
\newblock Multi-task domain adaptation for deep learning of instance grasping
  from simulation.
\newblock \emph{CoRR}, abs/1710.06422, 2017.

\bibitem[Bousmalis et~al.(2017)Bousmalis, Irpan, Wohlhart, Bai, Kelcey,
  Kalakrishnan, Downs, Ibarz, Pastor, Konolige, Levine, and
  Vanhoucke]{bousmalis2017using}
Konstantinos Bousmalis, Alex Irpan, Paul Wohlhart, Yunfei Bai, Matthew Kelcey,
  Mrinal Kalakrishnan, Laura Downs, Julian Ibarz, Peter Pastor, Kurt Konolige,
  Sergey Levine, and Vincent Vanhoucke.
\newblock Using simulation and domain adaptation to improve efficiency of deep
  robotic grasping.
\newblock \emph{CoRR}, abs/1709.07857, 2017.

\bibitem[Kenneally et~al.(2016)Kenneally, De, and Koditschek]{7403902}
Gavin Kenneally, Avik De, and Daniel~E Koditschek.
\newblock Design principles for a family of direct-drive legged robots.
\newblock \emph{IEEE Robotics and Automation Letters}, 1\penalty0 (2):\penalty0
  900--907, 2016.

\bibitem[Coumans and Bai(2016--2017)]{coumans2017}
Erwin Coumans and Yunfei Bai.
\newblock Pybullet, a python module for physics simulation in robotics, games
  and machine learning.
\newblock \url{http://pybullet.org}, 2016--2017.

\bibitem[Peng et~al.(2017{\natexlab{c}})Peng, van~de Panne, and
  Yin]{2017-SCA-action}
Xue~Bin Peng, Michiel van~de Panne, and KangKang Yin.
\newblock Learning locomotion skills using {DeepRL}: Does the choice of action
  space matter?
\newblock In \emph{Proc. ACM SIGGRAPH / Eurographics Symposium on Computer
  Animation}, 2017{\natexlab{c}}.

\bibitem[urd()]{urdf}
{URDF} - {ROS} wiki.
\newblock \url{http://wiki.ros.org/urdf}.

\bibitem[De and Koditschek(2015)]{DBLP:journals/corr/DeK15}
Avik De and Daniel~E. Koditschek.
\newblock The {P}enn {J}erboa: {A} platform for exploring parallel composition
  of templates.
\newblock \emph{CoRR}, abs/1502.05347, 2015.

\bibitem[Luo and Hauser(2017)]{luo2017robust}
Jingru Luo and Kris Hauser.
\newblock Robust trajectory optimization under frictional contact with
  iterative learning.
\newblock \emph{Autonomous Robots}, 41\penalty0 (6):\penalty0 1447--1461, 2017.

\bibitem[Hirai and Gunji(2000)]{Slipperiness}
Ikuko Hirai and Toshihiro Gunji.
\newblock Slipperiness and coefficient of friction on the carpets.
\newblock \emph{Sen'i Kikai Gakkaishi (Journal of the Textile Machinery Society
  of Japan)}, 53:\penalty0 T140--T146, 2000.

\end{thebibliography}

\end{document}